\newif\iftwocol\twocoltrue              

\documentclass[twoside,twocolumn]{article}
\usepackage{latexsym,amsmath,amssymb,hyperref}
\usepackage{wrapfig}
\def\eqvsp{}  \newdimen\paravsp  \paravsp=1.3ex
\makeatletter
\def\section{\@startsection {section}{1}{\z@}{-2.0ex plus
    -0.5ex minus -.2ex}{1.5ex plus 0.3ex minus .2ex}{\large\bf\raggedright}}
\makeatother
\def\,{\mskip 3mu} \def\>{\mskip 4mu plus 2mu minus 4mu} \def\;{\mskip 5mu plus 5mu} \def\!{\mskip-3mu}
\def\dispmuskip{\thinmuskip= 3mu plus 0mu minus 2mu \medmuskip=  4mu plus 2mu minus 2mu \thickmuskip=5mu plus 5mu minus 2mu}
\def\textmuskip{\thinmuskip= 0mu                    \medmuskip=  1mu plus 1mu minus 1mu \thickmuskip=2mu plus 3mu minus 1mu}
\textmuskip
\def\beq{\eqvsp\dispmuskip\begin{equation}}    \def\eeq{\eqvsp\end{equation}\textmuskip}
\def\beqn{\eqvsp\dispmuskip\begin{displaymath}}\def\eeqn{\eqvsp\end{displaymath}\textmuskip}
\def\bqa{\eqvsp\dispmuskip\begin{eqnarray}}    \def\eqa{\eqvsp\end{eqnarray}\textmuskip}
\def\bqan{\eqvsp\dispmuskip\begin{eqnarray*}}  \def\eqan{\eqvsp\end{eqnarray*}\textmuskip}
\def\paradot#1{\vspace{\paravsp plus 0.5\paravsp minus 0.5\paravsp}\noindent{\bf\boldmath{#1.}}}
\def\paranodot#1{\vspace{\paravsp plus 0.5\paravsp minus 0.5\paravsp}\noindent{\bf\boldmath{#1}}}
\def\req#1{(\ref{#1})}

\def\nq{\hspace{-1em}}

\def\fr#1#2{{\textstyle{#1\over#2}}}

\def\SetR{I\!\!R}
\def\SetN{I\!\!N}

\def\e{{\rm e}}                        

\def\P{{\rm P}}                         

\def\v{\boldsymbol}
\def\trp{{\!\top\!}}

\def\a{\alpha}

\def\s{\sigma}

\def\A{{\cal A}}
\def\O{{\cal O}}
\def\R{{\cal R}}
\def\S{{\cal S}}
\def\H{{\cal H}}

\def\T{{\cal T}}
\def\Agent{\text{Agent}}
\def\Env{\text{Env}}
\def\p{{\scriptscriptstyle+}}
\def\vi{{\scriptscriptstyle\bullet}}
\def\CL{\text{CL}}
\def\CLN{\text{CL}}
\def\Cost{\text{Cost}}
\def\Pa{\text{Pa}}

\begin{document}

\title{\vspace{-4ex}
\vskip 2mm\bf\Large\hrule height5pt \vskip 4mm
Feature Dynamic Bayesian Networks
\vskip 4mm \hrule height2pt}
\author{{\bf Marcus Hutter}\\[3mm]
\normalsize RSISE$\,$@$\,$ANU and SML$\,$@$\,$NICTA \\
\normalsize Canberra, ACT, 0200, Australia \\
\normalsize \texttt{marcus@hutter1.net \ \  www.hutter1.net}
}
\date{24 December 2008}
\maketitle

\begin{abstract}\bf
Feature Markov Decision Processes ($\Phi$MDPs)
\cite{Hutter:09phimdp} are well-suited for learning agents in
general environments. Nevertheless, unstructured ($\Phi$)MDPs are
limited to relatively simple environments. Structured MDPs like
Dynamic Bayesian Networks (DBNs) are used for large-scale real-world
problems. In this article I extend $\Phi$MDP to $\Phi$DBN. The
primary contribution is to derive a cost criterion that allows to
automatically extract the most relevant features from the
environment, leading to the ``best'' DBN representation. I discuss
all building blocks required for a complete general learning
algorithm.

{\it Keywords:}
Reinforcement learning;
dynamic Bayesian network;
structure learning;
feature learning;
global vs. local reward;
explore-exploit.
\end{abstract}

\section{Introduction}\label{secIntro}

\paradot{Agents}
The agent-environment setup in which an {\em Agent} interacts with
an {\em Environment} is a very general and prevalent framework for
studying intelligent learning systems \cite{Russell:03}.
In cycles $t=1,2,3,...$, the environment provides a (regular) {\em
observation} $o_t\in\O$ (e.g.\ a camera image) to the agent; then
the agent chooses an {\em action} $a_t\in\A$ (e.g.\ a limb
movement); finally the environment provides a real-valued {\em
reward} $r_t\in\SetR$ to the agent. The reward may be very scarce,
e.g.\ just +1 (-1) for winning (losing) a chess game, and 0 at all
other times \cite[Sec.6.3]{Hutter:04uaibook}. Then the next cycle
$t+1$ starts.
The agent's objective is to maximize his reward.

\paradot{Environments}
For example, {\em sequence prediction} is concerned with
environments that do not react to the agents actions (e.g.\ a
weather-forecasting ``action'') \cite{Hutter:03optisp}, {\em
planning} deals with the case where the environmental function is
known \cite{Ross:08pomdp}, {\em classification} and {\em regression}
is for conditionally independent observations \cite{Bishop:06}, {\em
Markov Decision Processes} (MDPs) assume that $o_t$ and $r_t$ only
depend on $a_{t-1}$ and $o_{t-1}$ \cite{Sutton:98}, POMDPs deal with
{\em Partially Observable MDPs} \cite{Kaelbling:98}, and {\em
Dynamic Bayesian Networks} (DBNs) with structured MDPs
\cite{Boutilier:99}.

\paradot{Feature MDPs \cite{Hutter:09phimdp}}
Concrete real-world problems can often be {\em modeled} as MDPs. For
this purpose, a {\em designer} extracts relevant features from the
history (e.g.\ position and velocity of all objects), i.e.\ the {\em
history} $h_t=a_1o_1r_1...a_{t-1}o_{t-1}r_{t-1}o_t$ is summarized by a
{\em feature} vector $s_t:=\Phi(h_t)$. The feature
vectors are regarded as {\em states} of an MDP and are assumed to be
(approximately) Markov.

Artificial General Intelligence (AGI) \cite{Goertzel:07} is
concerned with designing {\em agents that perform well in a very
large range of environments} \cite{Hutter:07iorx}, including all of
the mentioned ones above and more. In this general situation, it is
not a priori clear what the useful features are. Indeed, any
observation in the (far) past may be relevant in the future. A
solution suggested in \cite{Hutter:09phimdp} is to learn $\Phi$
itself.

If $\Phi$ keeps too much of the history (e.g.\ $\Phi(h_t)=h_t$), the
resulting MDP is too large (infinite) and cannot be learned. If
$\Phi$ keeps too little, the resulting state sequence is not Markov.
The {\em Cost} criterion I develop formalizes this tradeoff and is
minimized for the ``best'' $\Phi$.
At any time $n$, the best $\Phi$ is the one that minimizes the
Markov code length of $s_1...s_n$ and $r_1...r_n$. This reminds but
is actually quite different from MDL, which minimizes model+data
code length \cite{Gruenwald:07book}.

\paradot{Dynamic Bayesian networks}
The use of ``unstructured'' MDPs \cite{Hutter:09phimdp}, even our
$\Phi$-optimal ones, is clearly limited to relatively simple tasks.
Real-world problems are structured and can often be represented by
dynamic Bayesian networks (DBNs) with a reasonable number of nodes
\cite{Dean:89}. Bayesian networks in general and DBNs in particular
are powerful tools for modeling and solving complex real-world
problems. Advances in theory and increase in computation power
constantly broaden their range of applicability
\cite{Boutilier:99,Strehl:07}.

\paradot{Main contribution}
The primary contribution of this work is to extend the {\em $\Phi$
selection principle} developed in \cite{Hutter:09phimdp} for MDPs to
the conceptually much more demanding DBN case.
The major extra complications are approximating, learning and coding
the rewards, the dependence of the Cost criterion on the DBN
structure, learning the DBN structure, and how to store and find the
optimal value function and policy.

Although this article is self-contained, it is recommended to read
\cite{Hutter:09phimdp} first.

\section{Feature Dynamic Bayesian Networks ($\mathbf\Phi$DBN)}\label{secFDBN}

In this section I recapitulate the definition of $\Phi$MDP from
\cite{Hutter:09phimdp}, and adapt it to DBNs. While formally a
DBN is just a special case of an MDP, exploiting the additional
structure efficiently is a challenge.
%
For generic MDPs, typical algorithms should be polynomial and can at
best be linear in the number of states $|\S|$. For DBNs we want
algorithms that are polynomial in the number of features $m$.
Such DBNs have exponentially many states ($2^{O(m)}$), hence the
standard MDP algorithms are exponential, not polynomial, in $m$.
Deriving poly-time (and poly-space!) algorithms for DBNs by
exploiting the additional DBN structure is the challenge. The gain
is that we can handle exponentially large structured MDPs
efficiently.

\paradot{Notation}
Throughout this article, $\log$ denotes the binary logarithm, %
and $\delta_{x,y}=\delta_{xy}=1$ if $x=y$ and $0$ else is the Kronecker symbol. %
I generally omit separating commas if no confusion arises, in
particular in indices.
For any $z$ of suitable type (string,vector,set), I define
string $\v z = z_{1:l} = z_1...z_l$, %
sum $z_\p=\sum_j z_j$, union $z_*=\bigcup_j z_j$, and vector $\v z_\vi=(z_1,...,z_l)$, %
where $j$ ranges over the full range $\{1,...,l\}$
and $l=|z|$ is the length or dimension or size of $z$. %
$\hat z$ denotes an estimate of $z$.
The characteristic function $1\!\!1_B=1$ if $B$=true and $0$ else.
$\P(\cdot)$ denotes a probability over states and rewards or
parts thereof. I do not distinguish between random variables $Z$
and realizations $z$, and abbreviation $\P(z):=\P[Z=z]$ never
leads to confusion.
%
More specifically, $m\in\SetN$ denotes the number of features, %
$i\in\{1,...,m\}$ any feature, %
$n\in\SetN$ the current time, %
and $t\in\{1,...,n\}$ any time. %
Further, in order not to get distracted at several places I gloss
over initial conditions or special cases where inessential. Also
0$*$undefined=0$*$infinity:=0.

\paradot{$\Phi$MDP definition}
A $\Phi$MDP consists of a 7 tupel ($\O,\A,\R,\Agent,\Env,\Phi,\S$)
= (observation space, action space, reward space, agent,
environment, feature map, state space). Without much loss of
generality, I assume that $\A$ and $\O$ are finite and
$\R\subseteq\SetR$. Implicitly I assume $\A$ to be small, while $\O$
may be huge.

$\Agent$ and $\Env$ are a pair or triple of interlocking
functions of the history $\H:=(\O\times\A\times\R)^*\times\O$:
\bqan
  & & \nq\Env:\H\times\A\times\R\leadsto\O, \quad   o_n = \Env(h_{n-1}a_{n-1}r_{n-1}), \\
  & & \nq\!\!\!\!\!\Agent:\H\leadsto\A, \qquad\qquad\quad a_n = \Agent(h_n), \\
  & & \nq\Env:\H\times\A\leadsto\R, \qquad\quad       r_n = \Env(h_n a_n).
\eqan
where $\leadsto$ indicates that mappings $\to$ might be stochastic.
The informal goal of AI is to design an $\Agent()$ that achieves high (expected)
reward over the agent's lifetime in a large range of $\Env()$ironments.

The feature map $\Phi$ maps histories to states
\beqn
\iftwocol
  \Phi:\H\to\S,\quad s_t=\Phi(h_t),\quad h_t=oar_{1:t-1}o_t\in\H
\else
  \Phi:\H\to\S,\qquad s_t=\Phi(h_t),\qquad h_t=o_1 a_1 r_1...o_{t-1} a_{t-1} r_{t-1}o_t\in\H
\fi
\eeqn
The idea is that $\Phi$ shall extract the ``relevant'' aspects of
the history in the sense that ``compressed'' history $sar_{1:n}
\equiv s_1 a_1 r_1...s_n a_n r_n$ can well be described as a sample
from some MDP ($\S,\A,T,R$) = (state space, action space, transition
probability, reward function).

\paranodot{($\Phi$) Dynamic Bayesian Networks}
are structured ($\Phi$)MDPs. The state space is $\S=\{0,1\}^m$, and each
state $s\equiv\v x\equiv(x^1,...,x^m)\in\S$ is interpreted as a
feature vector $\v x=\v\Phi(h)$, where $x^i=\Phi^i(h)$ is the value
of the $i$th binary feature. In the following I will also refer to
$x^i$ as feature $i$, although strictly speaking it is its value.
Since non-binary features can be realized as a list of binary
features, I restrict myself to the latter.

Given $\v x_{t-1}=\v x$, I assume that the features
$(x_t^1,...,x_t^m)=\v x'$ at time $t$ are independent, and that each
$x'^i$ depends only on a subset of ``parent'' features $\v
u^i\subseteq\{x^1,...,x^m\}$, i.e.\ the transition matrix has the
structure
\beq\label{TDBN}
  T_{\v x\v x'}^a \!= \P(\v x_t=\v x'|\v x_{t-1}=\v x,a_{t-1}=a)
  = \iftwocol\smash{\prod_{i=1}^m}\else \prod_{i=1}^m \fi \P^a(x'^i|\v u^i)
\eeq
This defines our {\bf$\mathbf\Phi$DBN model}. It is just a $\Phi$MDP
with special $\S$ and $T$. Explaining $\Phi$DBN on an example is
easier than staying general.

\section{$\mathbf\Phi$DBN Example}\label{secExample}

Consider an instantiation of the simple vacuum world
\cite[Sec.3.6]{Russell:03}. There are two rooms, $A$ and $B$, and a
vacuum $R$obot that can observe whether the room he is in is $C$lean
or $D$irty; $M$ove to the other room, $S$uck, i.e.\ clean the room
he is in; or do $N$othing. After 3 days a room gets dirty again.
Every clean room gives a reward 1, but a moving or sucking robot
costs and hence reduces the reward by 1.
Hence $\O=\{A,B\}\times\{C,D\}$, $\A=\{N,S,M\}$, $\R=\{-1,0,1,2\}$,
and the dynamics Env() (possible histories) is clear from the above
description.

\paradot{Dynamics as a DBN}
We can model the dynamics by a DBN as follows: The state is modeled
by 3 features. Feature $R\in\{A,B\}$ stores in which room the robot
is, and feature $A/B\in\{0,1,2,3\}$ remembers (capped at 3) how long
ago the robot has cleaned room $A/B$ last time, hence
$\S=\{0,1,2,3\}\times\{A,B\}\times\{0,1,2,3\}$. The state/feature
transition is as follows:
\iftwocol
\bqan
  & & \nq\nq\mbox{if ($x^R\!\!=\!A$ and $a\!=\!S$) then $x'^A\!\!=\!0$ else $x'^A\!\!=\!\min\{x^A\!\!+\!1,3\}$}; \\
  & & \nq\nq\mbox{if ($x^R\!\!=\!B$ and $a\!=\!S$) then $x'^B\!\!=\!0$ else $x'^B\!\!=\!\min\{x^B\!\!+\!1,3\}$}; \\
  & & \nq\nq\mbox{if $a\!=\!M$ (if $x^R\!\!=\!B$ then $x'^R\!\!=\!A$ else $x'^R\!\!=\!B$) else $x'^R\!\!=\!x^R$};
\eqan
\else
\bqan
  & & \mbox{if ($x^R=A$ and $a=S$) then $x'^A=0$ else $x'^A=\min\{x^A+1,3\}$}; \\
  & & \mbox{if ($x^R=B$ and $a=S$) then $x'^B=0$ else $x'^B=\min\{x^B+1,3\}$}; \\
  & & \mbox{if $a=M$ (if $x^R=B$ then $x'^R=A$ else $x'^R=B$) else $x'^R=x^R$};
\eqan
\fi
A DBN can be viewed as a two-layer Bayesian network
\cite{Boutilier:99}. The dependency
structure of our example is depicted in the right diagram.

\begin{wrapfigure}{r}{20mm}
\unitlength=2.2ex
\linethickness{0.4pt}
\begin{picture}(7,6)(0,2) 
\thinlines
\put(1,10){\makebox(0,0)[ct]{$t-1$}}
\put(5,10){\makebox(0,0)[ct]{$t$}}
\thicklines
\put(1,8){\circle{2}}\put(1,8){\makebox(0,0)[cc]{$A$}}
\put(5,8){\circle{2}}\put(5,8){\makebox(0,0)[cc]{$A'$}}
\put(1,5){\circle{2}}\put(1,5){\makebox(0,0)[cc]{$R$}}
\put(5,5){\circle{2}}\put(5,5){\makebox(0,0)[cc]{$R'$}}
\put(1,2){\circle{2}}\put(1,2){\makebox(0,0)[cc]{$B$}}
\put(5,2){\circle{2}}\put(5,2){\makebox(0,0)[cc]{$B'$}}
\thinlines
\put(1,0){\makebox(0,0)[cb]{$x$}}
\put(5,0){\makebox(0,0)[cb]{$x'$}}
\multiput(2,2)(0,3){3}{\vector(1,0){2}}
\put(1.8,5.6){\vector(4,3){2.4}}
\put(1.8,4.4){\vector(4,-3){2.4}}
\end{picture}
\end{wrapfigure}
Each feature consists of a (left,right)-pair of nodes, and a node
$i\in\{1,2,3=m\}\widehat=\{A,R,B\}$ on the right is connected to all
and only the parent features $\v u^i$ on the left.
The reward is
\beqn
  r=1\!\!1_{x^A<3} + 1\!\!1_{x^B<3} - 1\!\!1_{a\neq N}
\eeqn
The features map $\v\Phi=(\Phi^A,\Phi^R,\Phi^B)$ can also be written
down explicitly. It depends on the actions and observations of the
last 3 time steps.

\paradot{Discussion}
Note that all nodes $x'^i$ can implicitly also depend on the chosen
action $a$.
The optimal policies are repetitions of action sequence $S,N,M$ or $S,M,N$.
One might think that binary features $x^{A/B}\in\{C,D\}$ are sufficient,
but this would result in a POMDP (Partially Observable MDP), since the
cleanness of room $A$ is not observed while the robot is in room
$B$. That is, $\v x'$ would not be a (probabilistic) function of $\v
x$ and $a$ alone.
The quaternary feature $x^A\in\{0,1,2,3\}$ can easily be converted
into two binary features, and similarly $x^B$.
The purely deterministic example can easily be made stochastic. For
instance, $S$ucking and $M$oving may fail with a certain
probability. Possible, but more complicated is to model a
probabilistic transition from $C$lean to $D$irty. In the randomized
versions the agent needs to use its observations.

\section{$\mathbf\Phi$DBN Coding and Evaluation}\label{secFDBNcode}

I now construct a code for $s_{1:n}$ given $a_{1:n}$, and for $r_{1:n}$
given $s_{1:n}$ and $a_{1:n}$, which is optimal (minimal) if
$s_{1:n}r_{1:n}$ given $a_{1:n}$ is sampled from some MDP. It
constitutes our cost function for $\v\Phi$ and is used to define the
$\v\Phi$ selection principle for DBNs. Compared to the MDP case,
reward coding is more complex, and there is an extra dependence on
the graphical structure of the DBN.

Recall \cite{Hutter:09phimdp} that a sequence $z_{1:n}$ with counts
$\v n=(n_1,...,n_m)$ can within an additive constant be coded in
\beq\label{iidCodeDBN}
  \CLN(\v n) :=
  n\,H(\v n/n) + \fr{m'-1}2\log n
  \ \mbox{ if } \ n\!>\!0 \ \mbox{ and } \ 0 \ \mbox{ else}
\eeq
bits, where $n=n_+=n_1+...+n_m$ and $m'=|\{i:n_i>0\}|\leq m$ is the
number of non-empty categories, and $H(\v p):=-\sum_{i=1}^{m}p_i\log
p_i$ is the entropy of probability distribution $\v p$. The code is
optimal (within $+O(1)$) for all i.i.d.\ sources.

\paradot{State/Feature Coding}
Similarly to the $\Phi$MDP case, we need to code the temporal
``observed'' state=feature sequence $\v x_{1:n}$. I do this by a
frequency estimate of the state/feature transition probability.
(Within an additive constant, MDL, MML, combinatorial, incremental,
and Bayesian coding all lead to the same result). In the following I
will drop the prime in $(\v u^i,a,x'^i)$ tuples and related
situations if/since it does not lead to confusion.
Let $\T_{\v u^ix^i}^{ia} = \{t\leq n:\v u_{t-1}=\v u^i, a_{t-1}=a,
x_t^i=x^i\}$ be the set of times $t-1$ at which features that
influence $x^i$ have values $\v u^i$, and action is $a$, and which
leads to feature $i$ having value $x^i$. Let $n_{\v
u^ix^i}^{ia}=|\T_{\v u^ix^i}^{ia}|$ their number ($n_{\p\p}^{i\p}=n$
$\forall i$). I estimate each feature probability separately by
$\hat\P^a(x^i|\v u^i) = n_{\v u^ix^i}^{ia}/n_{\v u^i\p}^{ia}$.
Using \req{TDBN}, this yields
\iftwocol
\bqan
  \hat\P(\v x_{1:n}|a_{1:n})
  &=& \prod_{t=1}^n \hat T_{\v x_{t-1}\v x_t}^{a_{t-1}}
  \;=\; \prod_{t=1}^n \prod_{i=1}^m\hat\P^{a_{t-1}}(x_t^i|\v u_{t-1}^i)
\\ \vspace{-2ex}
  =\;\;\; ... &=& \exp\bigg[\sum_{i,\v u^i,a} n_{\v u^i\p}^{ia} H\bigg({\v n_{\v u^i\vi}^{ia}\over n_{\v u^i\p}^{ia}}\bigg)\bigg]
\eqan
\else
\beqn
  \hat\P(\v x_{1:n}|a_{1:n})
  \;=\; \prod_{t=1}^n \hat T_{\v x_{t-1}\v x_t}^{a_{t-1}}
  \;=\; \prod_{t=1}^n \prod_{i=1}^m\hat\P^{a_{t-1}}(x_t^i|\v u_{t-1}^i)
  \;=\; \exp\bigg[\sum_{i,\v u^i,a} n_{\v u^i\p}^{ia} H\bigg({\v n_{\v u^i\vi}^{ia}\over n_{\v u^i\p}^{ia}}\bigg)\bigg]
\eeqn
\fi
The length of the Shannon-Fano code of $\v x_{1:n}$ is just the
logarithm of this expression.
We also need to code each non-zero count $n_{\v u^ix^i}^{ia}$
to accuracy $O(1/\sqrt{n_{\smash{\;\v u^i\!\!\p}}^{ia}})$, which each needs
$\fr12\log(n_{\v u^i\p}^{ia})$ bits.
Together this gives
a complete code of length
\beq\label{xCodeDBN}
  \CL(\v x_{1:n}|a_{1:n}) \;=\; \sum_{i,\v u^i,a} \CLN(\v n_{\v u^i\vi}^{ia})
\eeq
The rewards are more complicated.

\paradot{Reward structure}
Let $R_{\v x\v x'}^a$ be (a model of) the observed reward when
action $a$ in state $\v x$ results in state $\v x'$. It is natural
to assume that the structure of the rewards $R_{\v x\v x'}^a$ is
related to the transition structure $T_{\v x\v x'}^a$. Indeed, this
is not restrictive, since one can always consider a DBN with the
union of transition and reward dependencies.
Usually it is assumed that the ``global'' reward is a {\em sum} of
``local'' rewards $R_{\v u^ix'^i}^{ia}$, one for each feature $i$
\cite{Koller:99}.
For simplicity of exposition I assume that the local reward $R^i$
only depends on the feature value $x'^i$ and not on $\v u^i$ and
$a$. Even this is not restrictive and actually may be advantageous
as discussed in \cite{Hutter:09phimdp} for MDPs. So I assume
\beqn
  R_{\v x\v x'}^a \;=\; \sum_{i=1}^m R_{x'^i}^i \;=:\; R(\v x')
\eeqn
For instance, in the example of Section \ref{secFDBN}, two local
rewards ($R_{x'^A}^A=1\!\!1_{x'^A<3}$ and
$R_{x'^B}^B=1\!\!1_{x'^B<3}$) depend on $\v x'$ only, but the third
reward depends on the action ($R^R=-1\!\!1_{a\neq N}$).

Often it is assumed that the local rewards are directly observed or
known \cite{Koller:99}, but we neither want nor can do this here:
Having to specify many local rewards is an extra burden for the
environment (e.g.\ the teacher), which preferably should be avoided.
In our case, it is not even possible to pre-specify a local reward
for each feature, since the features $\Phi^i$ themselves are learned
by the agent and are not statically available.
They are agent-internal and not part of the $\Phi$DBN interface.
In case multiple rewards {\em are} available, they can be modeled
as part of the regular observations $o$, and $r$ only holds the
overall reward. The agent must and can learn to interpret and
exploit the local rewards in $o$ by himself.

\paradot{Learning the reward function}
In analogy to the MDP case for $R$ 
and the DBN case for $T$ above it is tempting to estimate
$R_{x^i}^i$ by
$\sum_{r'}r' n_{\!\p x^i}^{ir'} / n_{\!\p x^i}^{i\p}$
but this makes no sense. For instance if
$r_t=1\,\forall t$, then $\hat R_{x^i}^i\equiv 1$, and $\hat R_{\v
x\v x'}^a\equiv m$ is a gross mis-estimation of $r_t\equiv 1$. The
localization of the global reward is somewhat more complicated. The
goal is to choose $R_{x^1}^1,...,R_{x^m}^m$ such that $r_t=R(\v
x_t)\,\forall t$.

Without loss we can set $R_0^i\equiv 0$, since we can subtract a
constant from each local reward and absorb them into an overall
constant $w_0$. This allows us to write
\beqn
  R(\v x) \;=\; w_0 x^0+w_1 x^1+...+w_m x^m \;=\; \v w^\trp\v x
\eeqn
where $w_i:=R_1^i$ and $x^0:\equiv 1$.

In practice, the $\Phi$DBN model will not be perfect, and an
approximate solution, e.g.\ a least squares fit, is the best we can
achieve. The square loss can be written as
\beq\label{RLoss}
  \mbox{Loss}(\v w) \;:=\; \sum_{t=1}^n (R(\v x_t)-r_t)^2
  \;=\; \v w^\trp A\v w - 2\v b^\trp\v w + c
\eeq\vspace{-1ex}
\beqn
  A_{ij}:=\sum_{t=1}^n x^i_t x^j_t,\quad
  b_i:=\sum_{t=1}^n r_t x^i_t,\quad
  c:=\sum_{t=1}^n r_t^2
\eeqn
Note that $A_{ij}$ counts the number of times feature $i$ and $j$
are ``on'' (=1) simultaneously, and $b_i$ sums all rewards for which
feature $i$ is on. The loss is minimized for
\beqn
  \v{\hat w} := \arg\min_{\v w}\mbox{Loss}(\v w) = A^{-1}\v b,\qquad
  \hat R(\v x)= \v{\hat w}^\trp\v x
\eeqn
which involves an inversion of the $(m+1)\times(m+1)$ matrix $A$.
For singular $A$ we take the pseudo-inverse.

\paradot{Reward coding}
The quadratic loss function suggests a Gaussian model for the
rewards:
\beqn
  \P(r_{1:n}|\v{\hat w},\s) \;:=\;
  {\exp(-\mbox{Loss}(\v{\hat w})/2\s^2)/(2\pi\s^2)^{n/2}}
\eeqn
Maximizing this w.r.t.\ the variance $\sigma^2$ yields the maximum
likelihood estimate
\beqn
  -\log\P(r_{1:n}|\v{\hat w},\hat\s) \;=\; \fr n2\log(\mbox{Loss}(\v{\hat w})) - \fr n2\log\fr{n\e}{2\pi}
\eeqn
where $\hat\s^2=\mbox{Loss}(\v{\hat w})/n$. Given $\v{\hat w}$ and
$\hat\s$ this can be regarded as the (Shannon-Fano) code length of
$r_{1:n}$ (there are actually a few subtleties here which I gloss
over). Each weight $\hat w_k$ and $\hat\s$ need also be
coded to accuracy $O(1/\sqrt{n})$, which needs $(m+2)\fr12\log n$
bits total. Together this gives a complete code of length
\beq\label{rCodeDBN}
  \CL(r_{1:n}|\v x_{1:n}a_{1:n})
\iftwocol \;=\; \hspace*{21.5ex}\eeq\beqn\fi
  \;=\; \fr{n}{2}\log(\mbox{Loss}(\v{\hat w})) + \fr{m+2}{2}\log n - \fr n2\log\fr{n\e}{2\pi}
\iftwocol\eeqn\else\eeq\fi

\paranodot{$\mathbf\Phi$DBN evaluation and selection}
is similar to the MDP case. Let $G$ denote the graphical structure
of the DBN, i.e.\ the set of parents $\Pa^i\subseteq\{1,...,m\}$ of
each feature $i$. (Remember $\v u^i$ are the parent values).
Similarly to the MDP case, the cost of
$(\v\Phi,G)$ on $h_n$ is defined as
\beq\label{costPhiG}
  \Cost(\v\Phi,G|h_n) \;:=\;
  \CL(\v x_{1:n}|a_{1:n}) + \CL(r_{1:n}|\v x_{1:n},a_{1:n}),
\eeq
and the best $(\v\Phi,G)$ minimizes this cost.
\beqn\label{bestPhiG}
  (\v\Phi^{best},G^{best}) \;:=\; \arg\min_{\v\Phi,G}\{\Cost(\v\Phi,G|h_n)\}
\eeqn
A general discussion why this is a good criterion can be found in
\cite{Hutter:09phimdp}.
In the following section I mainly highlight the difference to the
MDP case, in particular the additional dependence on and
optimization over $G$.

\section{DBN Structure Learning \& Updating}\label{secDBNSLU}

This section briefly discusses minimization of \req{costPhiG}
w.r.t.\ $G$ given $\v\Phi$ and even briefer minimization w.r.t.\
$\v\Phi$. For the moment regard $\v\Phi$ as given and fixed.

\paradot{Cost and DBN structure}
For general structured local rewards $R_{\v u^ix'^i}^{ia}$,
\req{xCodeDBN} and \req{rCodeDBN} both depend on $G$, and
\req{costPhiG} represents a novel DBN structure learning criterion
that includes the rewards.

For our simple reward model $R_{x^i}^i$, \req{rCodeDBN} is
independent of $G$, hence only \req{xCodeDBN} needs to be
considered. This is a standard MDL criterion, but I have not seen it
used in DBNs before. Further, the features $i$ are independent in
the sense that we can search for the optimal parent sets
$\Pa^i\subseteq\{1,...,m\}$ for each feature $i$ separately.

\paradot{Complexity of structure search}
Even in this case, finding the optimal DBN structure is generally
hard. In principle we could rely on off-the-shelf heuristic search
methods for finding good $G$, but it is probably better to use or
develop some special purpose optimizer. One may even restrict the
space of considered graphs $G$ to those for which \req{costPhiG} can
be minimized w.r.t.\ $G$ efficiently, as long as this restriction
can be compensated by ``smarter'' $\v\Phi$.

A brute force exhaustive search algorithm for $\Pa^i$ is to consider
all $2^m$ subsets of $\{1,...,m\}$ and select the one that minimizes
$\sum_{\v u^i,a} \CLN(\v n_{\v u^i\vi}^{ia})$. A reasonable and
often employed assumption is to limit the number of parents to some
small value $p$, which reduces the search space size to $O(m^p)$.

Indeed, since the Cost is exponential in the maximal number of parents
of a feature, but only linear in $n$, a Cost minimizing $\Phi$ can
usually not have more than a logarithmic number of parents,
which leads to a search space that is pseudo-polynomial in $m$.

\paradot{Heuristic structure search}
We could also replace the well-founded criterion \req{xCodeDBN} by
some heuristic. One such heuristic has been developed in
\cite{Strehl:07}. The mutual information is another popular
criterion for determining the dependency of two random variables, so
we could add $j$ as a parent of feature $i$ if the mutual
information of $x^j$ and $x'^i$ is above a certain threshold.
Overall this takes time $O(m^2)$ to determine $G$. An MDL inspired
threshold for binary random variables is $\fr1{2n}\log n$. Since the
mutual information treats parents independently, $\hat T$ has to be
estimated accordingly, essentially as in naive Bayes classification
\cite{Lewis:98} with feature selection, where $x'^i$ represents the
class label and $\v u^i$ are the features selected $\v x$. The
improved Tree-Augmented naive Bayes (TAN) classifier
\cite{Friedman:97} could be used to model synchronous feature
dependencies (i.e.\ within a time slice). The Chow-Liu
\cite{Chow:68} minimum spanning tree algorithm allows determining
$G$ in time $O(m^3)$. A tree becomes a forest if we employ a lower
threshold for the mutual information.

\paranodot{$\v\Phi$ search}
is even harder than structure search, and remains an art.
Nevertheless the reduction of the complex (ill-defined)
reinforcement learning problem to an internal feature search problem
with well-defined objective is a clear conceptual advance.

In principle (but not in practice) we could consider the set of {\em
all} (computable) functions $\{\Phi:\H\to\{0,1\}\}$. We then compute
$\Cost(\v\Phi|h)$ for every finite subset
$\v\Phi=\{\Phi^{i_1},...,\Phi^{i_m}\}$ and take the minimum (note
that the order is irrelevant).

Most practical search algorithms require the specification of some
neighborhood function, here for $\v\Phi$. For instance, stochastic
search algorithms suggest and accept a neighbor of $\v\Phi$ with a
probability that depends on the Cost reduction. See
\cite{Hutter:09phimdp} for more details. Here I will only present
some very simplistic ideas for features and neighborhoods.

Assume binary observations $\O=\{0,1\}$ and consider the last $m$
observations as features, i.e.\ $\Phi^i(h_n)=o_{n-i+1}$ and
$\v\Phi(h_n)=(\Phi^1(h_n),...,\Phi^m(h_n))=o_{n-m+1:n}$. So the
states are the same as for $\Phi_m$MDP in \cite{Hutter:09phimdp},
but now $\S=\{0,1\}^m$ is structured as $m$ binary features. In the
example here, $m=5$ lead to a perfect $\Phi$DBN. We can add a new
feature $o_{n-m}$ ($m\leadsto m+1$) or remove the last feature
($m\leadsto m-1$), which defines a natural neighborhood structure.

Note that the context trees of \cite{McCallum:96,Hutter:09phimdp}
are more flexible. To achieve this flexibility here we either have
to use smarter features within our framework (simply interpret
$s=\Phi_\S(h)$ as a feature vector of length
$m=\lceil\log|\S|\rceil$) or use smarter (non-tabular) estimates of
$\P^a(x^i|\v u^i)$ extending our framework (to tree dependencies).

For general purpose intelligent agents we clearly need more powerful
features. Logical expressions or (non)accepting Turing machines or
recursive sets can map histories or parts thereof into true/false or
accept/reject or in/out, respectively, hence naturally represent
binary features. Randomly generating such expressions or programs
with an appropriate bias towards simple ones is a universal feature
generator that eventually finds the optimal feature map. The idea is
known as Universal Search \cite{Gaglio:07}.

\section{Value \& Policy Learning in $\mathbf\Phi$DBN}\label{secVPLDBN}

Given an estimate $\hat{\v\Phi}$ of ${\v\Phi}^{best}$, the next step is to
determine a good action for our agent. I mainly concentrate on the
difficulties one faces in adapting MDP algorithms and discuss state
of the art DBN algorithms.
%
Value and policy learning in known finite state MDPs is easy
provided one is satisfied with a polynomial time algorithm.
Since a DBN is just a special (structured) MDP, its ($Q$) $V\!$alue
function respects the same Bellman equations
\cite[Eq.(6)]{Hutter:09phimdp}, 
and the optimal policy is
still given by $a_{n+1} := \arg\max_a Q_{\v x_{n+1}}^{*a}$. 
Nevertheless, their solution is now a nightmare, since the state
space is exponential in the number of features. We need algorithms
that are polynomial in the number of features, i.e.\ logarithmic in
the number of states.

\paradot{Value function approximation}
The first problem is that the optimal value and policy do not
respect the structure of the DBN. They are usually complex functions
of the (exponentially many) states, which cannot even be stored, not
to mention computed \cite{Koller:99}.
It has been suggested that the value can often be approximated well
as a sum of local values similarly to the rewards. Such a value
function can at least be stored.

\paradot{Model-based learning}
The default quality measure for the approximate value is the
$\rho$-weighted squared difference, where $\rho$ is the stationary
distribution.

Even for a fixed policy, value iteration does {\em not} converge to
the best approximation, but usually converges to a fixed point close
to it \cite{Bertsekas:96}.
Value iteration requires $\rho$ explicitly. Since $\rho$ is also too
large to store, one has to approximate $\rho$ as well.
Another problem, as pointed out in \cite{Koller:00}, is that
policy iteration may not converge, since different policies
have different (misleading) stationary distributions.
Koller and Parr \cite{Koller:00} devised algorithms for
general factored $\rho$, and Guestrin et al.\ \cite{Guestrin:03}
for max-norm, alleviating this problem.
Finally, general policies cannot be stored exactly, and another
restriction or approximation is necessary.

\paradot{Model-free learning}
Given the difficulties above, I suggest to (re)consider a very
simple class of algorithms, without suggesting that it is better.
The above model-based algorithms exploit $\hat T$ and $\hat R$
directly. An alternative is to sample from $\hat T$ and use model-free
``Temporal Difference (TD)'' learning algorithms based only on this
internal virtual sample \cite{Sutton:98}. We could use TD($\lambda)$
or $Q$-value variants with linear value function approximation.

Beside their simplicity, another advantage is that neither the
stationary distribution nor the policy needs to be stored or
approximated. Once approximation $\hat Q^*$ has been obtained, it is
trivial to determine the optimal (w.r.t.\ $\hat Q^*$) action via
$a_{n+1} = \arg\max_a Q_{\v x_{n+1}}^{*a}$ 
for any state of interest (namely $\v x_{n+1}$) exactly.

\paradot{Exploration}
Optimal actions based on approximate rather than exact values can
lead to very poor behavior due to lack of exploration. There are
polynomially optimal algorithms (Rmax,E3,OIM) for the
exploration-exploitation dilemma.

For model-based learning, extending E3 to DBNs is straightforward,
but E3 needs an oracle for planning in a given DBN
\cite{Kearns:99rl}. Recently, Strehl et al.\ \cite{Strehl:07}
accomplished the same for Rmax. They even learn the DBN structure,
albeit in a very simplistic way. Algorithm OIM \cite{Szita:08},
which I described in \cite{Hutter:09phimdp} for MDPs, can also likely be
generalized to DBNs, and I can imagine a model-free version.

\section{Incremental Updates}\label{secIU}

As discussed in Section \ref{secDBNSLU}, most search algorithms are
local in the sense that they produce a chain of ``slightly''
modified candidate solutions, here $\v\Phi$'s. This suggests a
potential speedup by computing quantities of interest incrementally.

\paradot{Cost}
Computing $\CL(\v x|\v a)$ in \req{xCodeDBN} takes at most time $O(m
2^k|\A|)$, where $k$ is the maximal number of parents of a feature.
If we remove feature $i$, we can simply remove/subtract the
contributions from $i$ in the sum. If we add a new feature $m+1$, we
only need to search for the best parent set $\v u^{m+1}$ for this
new feature, and add the corresponding code length. In practice,
many transitions don't occur, i.e.\ $n_{\v u^i x^i}^{ia}=0$, so
$\CL(\v x|\v a)$ can actually be computed much faster in time
$O(|\{n_{\v u^i x^i}^{ia}>0\}|)$, and incrementally even faster.

\paradot{Rewards}
When adding a new feature, the current local reward estimates may
not change much. If we reassign a fraction $\a\leq 1$ of reward to
the new feature $x^{m+1}$, we get the following ansatz\footnote{An
{\em Ansatz} is an initial mathematical or physical model with some
free parameters to be determined subsequently.
[http://en.wikipedia.org/wiki/Ansatz]}.
\bqan
  & & \nq\nq\hat R(x^1\!\!,...,x^{m+1})
  = (1\!-\!\a)\hat R(\v x) \!+\! w_{m+1}x^{m+1}
  =: \v v^\trp\v\psi(\v x)
\\
  & & \nq \v v \;:=\; (1\!-\!\a,w_{m+1})^\trp,\quad
        \v\psi \;:=\; (\hat R(\v x),x^{m+1})^\trp
\eqan
Minimizing $\sum_{t=1}^n (\hat R(x^1_t...x^{m+1}_t)-r_t)^2$ w.r.t.\
$\v v$ analogous to \req{RLoss} just requires a trivial $2\times 2$
matrix inversion. The minimum $\tilde{\v v}$ results in an initial
new estimate $\v{\tilde w}=((1-\tilde\a)\hat
w_0,...,(1-\tilde\a)\hat w_m,\tilde w_{m+1})^\trp$, which can be improved by
some first order gradient decent algorithm in time $O(m)$, compared
to the exact $O(m^3)$ algorithm.
When removing a feature, we simply redistribute its local reward to
the other features, e.g.\ uniformly, followed by improvement
steps that cost $O(m)$ time.

\paradot{Value}
All iteration algorithms described in Section \ref{secVPLDBN} for
computing ($Q$) $V\!$alues need an initial value for $V$ or $Q$. We
can take the estimate $\hat V$ from a previous $\v\Phi$ as an
initial value for the new $\v\Phi$. Similarly as for the rewards, we
can redistribute a fraction of the values by solving relatively
small systems of equations. The result is then used as an initial
value for the iteration algorithms in Section \ref{secVPLDBN}. A
further speedup can be obtained by using prioritized iteration
algorithms that concentrate their time on badly estimated
parameters, which are in our case the new values \cite{Sutton:98}.

Similarly, results from time $t$ can be (re)used as initial estimates
for the next cycle $t+1$, followed by a fast improvement step.

\section{Outlook}\label{secDisc}

$\Phi$DBN leaves much more questions open and room for modifications
and improvements than $\Phi$MDP. Here are a few.

\begin{itemize}\parskip=0ex\parsep=0ex\itemsep=0ex
\item The cost function can be improved by integrating out the states
analogous to the $\Phi$MDP case \cite{Hutter:09phimdp}: The
likelihood $\P(r_{1:n}|a_{1:n},\hat U)$ is unchanged, except that
$\hat U\equiv \hat T\hat R$ is now estimated locally, and the
complexity penalty becomes $\fr12(M+m+2)\log n$, where $M$ is
(essentially) the number of non-zero counts $n_{\v u^i x^i}^{ia}$,
but an efficient algorithm has yet to be found.

\item It may be necessary to impose and exploit structure on the
conditional probability tables $P^a(x^i|\v u^i)$ themselves
\cite{Boutilier:99}.

\item Real-valued observations and beliefs suggest to
extend the binary feature model to $[0,1]$ interval valued features
rather than coding them binary.
Since any continuous semantics that preserves the role of 0 and 1
is acceptable, there should be an efficient way to generalize Cost and
$V\!$alue estimation procedures.

\item I assumed that the reward/value is linear in local rewards/values.
Is this sufficient for all practical purposes?
I also assumed a least squares and Gaussian model for the local
rewards. There are efficient algorithms for much more flexible
models. The least we could do is to code w.r.t.\ the proper
covariance $A$.

\item I also barely discussed synchronous (within time-slice) dependencies.

\item I guess $\Phi$DBN will often be able to work around too restrictive
DBN models, by finding features $\Phi$ that are more compatible with
the DBN and reward structure.

\item Extra edges in the DBN can improve the linear value function
approximation. To give $\Phi$DBN incentives to do so,
the $V\!$alue would have to be included in the Cost criterion.

\item Implicitly I assumed that the action space $\A$ is small.
It is possible to extend $\Phi$DBN to large structured action spaces.

\item Apart from the $\Phi$-search, all parts of $\Phi$DBN seem to be
poly-time approximable, which is satisfactory in theory. In
practice, this needs to be improved to essentially linear time in
$n$ and $m$.

\item Developing smart $\Phi$ generation and smart stochastic search
algorithms for $\Phi$ are the major open challenges.

\item A more Bayesian Cost criterion
would be desirable: a likelihood of $h$ given $\Phi$ and a prior
over $\Phi$ leading to a posterior of $\Phi$ given $h$, or so. Monte
Carlo (search) algorithms like Metropolis-Hastings could sample from such a
posterior. Currently probabilities ($\widehat= 2^{-\CL}$) are
assigned only to rewards and states, but not to observations and
feature maps.
\end{itemize}

\paradot{Summary}
In this work I introduced a powerful framework ($\Phi$DBN) for
general-purpose intelligent learning agents, and presented
algorithms for all required building blocks.
The introduced cost criterion reduced the informal reinforcement
learning problem to an internal well-defined search for ``relevant''
features.


\begin{small}

\end{small}


\begin{thebibliography}{ABCD}

\bibitem[BDH99]{Boutilier:99}
C.~Boutilier, T.~Dean, and S.~Hanks.
\newblock Decision-theoretic planning: Structural assumptions and computational
  leverage.
\newblock {\em Journal of Artificial Intelligence Research}, 11:1--94, 1999.

\bibitem[Bis06]{Bishop:06}
C.~M. Bishop.
\newblock {\em Pattern Recognition and Machine Learning}.
\newblock Springer, 2006.

\bibitem[BT96]{Bertsekas:96}
D.~P. Bertsekas and J.~N. Tsitsiklis.
\newblock {\em Neuro-Dynamic Programming}.
\newblock Athena Scientific, Belmont, MA, 1996.

\bibitem[CL68]{Chow:68}
C.~K. Chow and C.~N. Liu.
\newblock Approximating discrete probability distributions with dependence
  trees.
\newblock {\em {IEEE} Transactions on Information Theory}, IT-14(3):462--467,
  1968.

\bibitem[DK89]{Dean:89}
T.~Dean and K.~Kanazawa.
\newblock A model for reasoning about persistence and causation.
\newblock {\em Computational Intelligence}, 5(3):142--150, 1989.

\bibitem[FGG97]{Friedman:97}
N.~Friedman, D.~Geiger, and M.~Goldszmid.
\newblock Bayesian network classifiers.
\newblock {\em Machine Learning}, 29(2):131--163, 1997.

\bibitem[Gag07]{Gaglio:07}
M.~Gaglio.
\newblock Universal search.
\newblock {\em Scholarpedia}, 2(11):2575, 2007.

\bibitem[GKPV03]{Guestrin:03}
C.~Guestrin, D.~Koller, R.~Parr, and S.~Venkataraman.
\newblock Efficient solution algorithms for factored {MDP}s.
\newblock {\em Journal of Artificial Intelligence Research (JAIR)},
  19:399--468, 2003.

\bibitem[GP07]{Goertzel:07}
B.~Goertzel and C.~Pennachin, editors.
\newblock {\em Artificial General Intelligence}.
\newblock Springer, 2007.

\bibitem[Gr{\"u}07]{Gruenwald:07book}
P.~D. Gr{\"u}nwald.
\newblock {\em The Minimum Description Length Principle}.
\newblock The MIT Press, Cambridge, 2007.

\bibitem[Hut03]{Hutter:03optisp}
M.~Hutter.
\newblock Optimality of universal {B}ayesian prediction for general loss and
  alphabet.
\newblock {\em Journal of Machine Learning Research}, 4:971--1000, 2003.

\bibitem[Hut05]{Hutter:04uaibook}
M.~Hutter.
\newblock {\em Universal Artificial Intelligence: Sequential Decisions based on
  Algorithmic Probability}.
\newblock Springer, Berlin, 2005.
\newblock 300 pages, http://www.hutter1.net/ai/uaibook.htm.

\bibitem[Hut09]{Hutter:09phimdp}
M.~Hutter.
\newblock Feature {M}arkov decision processes.
\newblock In {\em Artificial General Intelligence ({AGI'09})}. Atlantis Press,
  2009.

\bibitem[KK99]{Kearns:99rl}
M.~Kearns and D.~Koller.
\newblock Efficient reinforcement learning in factored {MDPs}.
\newblock In {\em Proc. 16th International Joint Conference on Artificial
  Intelligence ({IJCAI}-99)}, pages 740--747, San Francisco, 1999. Morgan
  Kaufmann.

\bibitem[KLC98]{Kaelbling:98}
L.~P. Kaelbling, M.~L. Littman, and A.~R. Cassandra.
\newblock Planning and acting in partially observable stochastic domains.
\newblock {\em Artificial Intelligence}, 101:99--134, 1998.

\bibitem[KP99]{Koller:99}
D.~Koller and R.~Parr.
\newblock Computing factored value functions for policies in structured
  {MDP}s,.
\newblock In {\em Proc. 16st International Joint Conf. on Artificial
  Intelligence ({IJCAI'99})}, pages 1332--1339, Edinburgh, 1999.

\bibitem[KP00]{Koller:00}
D.~Koller and R.~Parr.
\newblock Policy iteration for factored {MDPs}.
\newblock In {\em Proc. 16th Conference on Uncertainty in Artificial
  Intelligence ({UAI}-00)}, pages 326--334, San Francisco, CA, 2000. Morgan
  Kaufmann.

\bibitem[Lew98]{Lewis:98}
D.~D. Lewis.
\newblock Naive ({B}ayes) at forty: The independence assumption in information
  retrieval.
\newblock In {\em Proc. 10th European Conference on Machine Learning
  ({ECML'98})}, pages 4--15, Chemnitz, DE, 1998. Springer.

\bibitem[LH07]{Hutter:07iorx}
S.~Legg and M.~Hutter.
\newblock Universal intelligence: A definition of machine intelligence.
\newblock {\em Minds \& Machines}, 17(4):391--444, 2007.

\bibitem[McC96]{McCallum:96}
A.~K. McCallum.
\newblock {\em Reinforcement Learning with Selective Perception and Hidden
  State}.
\newblock PhD thesis, Department of Computer Science, University of Rochester,
  1996.

\bibitem[RN03]{Russell:03}
S.~J. Russell and P.~Norvig.
\newblock {\em Artificial Intelligence. {A} Modern Approach}.
\newblock Prentice-Hall, Englewood Cliffs, NJ, 2nd edition, 2003.

\bibitem[RPPCd08]{Ross:08pomdp}
S.~Ross, J.~Pineau, S.~Paquet, and B.~Chaib-draa.
\newblock Online planning algorithms for {POMDP}s.
\newblock {\em Journal of Artificial Intelligence Research}, 2008(32):663--704,
  2008.

\bibitem[SB98]{Sutton:98}
R.~S. Sutton and A.~G. Barto.
\newblock {\em Reinforcement Learning: An Introduction}.
\newblock MIT Press, Cambridge, MA, 1998.

\bibitem[SDL07]{Strehl:07}
A.~L. Strehl, C.~Diuk, and M.~L. Littman.
\newblock Efficient structure learning in factored-state {MDP}s.
\newblock In {\em Proc. 27th AAAI Conference on Artificial Intelligence}, pages
  645--650, Vancouver, BC, 2007. AAAI Press.

\bibitem[SL08]{Szita:08}
I.~Szita and A.~L{\"o}rincz.
\newblock The many faces of optimism: a unifying approach.
\newblock In {\em Proc. 12th International Conference ({ICML} 2008)}, volume
  307, Helsinki, Finland, June 2008.

\end{thebibliography}
\end{document}